\documentclass[conference,a4paper]{IEEEtran}

\usepackage{amsmath, amssymb, bm, color, graphicx, latexsym, amsthm}
\newtheorem{theorem}{Theorem}

\newcommand{\Fref}[1]{Fig.~\ref{#1}}
\newcommand{\figwidth}{8cm}

\begin{document}

\sloppy

\title{Sample Complexity of \\ Bayesian Optimal Dictionary Learning}

\author{
  \IEEEauthorblockN{Ayaka Sakata and Yoshiyuki Kabashima }
  \IEEEauthorblockA{Dep. of Computational Intelligence \& Systems Science\\
    Tokyo Institute of Technology \\
    Yokohama {226-8502}, Japan\\
    Email: ayaka@sp.dis.titech.ac.jp, kaba@dis.titech.ac.jp}
}



\maketitle

\begin{abstract}
We consider a learning problem of identifying a dictionary
matrix $\bm{D}\in \mathbb{R}^{M \times N}$ from
a sample set of $M$ dimensional vectors $\bm{Y}\in \mathbb{R}^{M \times P} = N^{-1/2} \bm{D} \bm{X}
\in \mathbb{R}^{M \times P}$, where $\bm{X} \in \mathbb{R}^{N \times P}$ is a sparse matrix in which
the density of non-zero entries is $0<\rho < 1$.
In particular, we focus on the minimum sample size $P_{\rm c}$
(sample complexity) necessary for perfectly identifying $\bm{D}$ of the optimal learning scheme when $\bm{D}$ and $\bm{X}$ are independently generated from
certain distributions. By using the replica method of statistical mechanics, we show that
$P_{\rm c} \sim O(N)$ holds as long as $\alpha =M/N >\rho$ is satisfied in the limit of $N\to\infty$.
Our analysis also implies that the posterior distribution given $\bm{Y}$ is
condensed only at the correct dictionary $\bm{D}$
when the compression rate
$\alpha$ is greater than a certain critical value $\alpha_{\rm M}(\rho)$.
This suggests that belief propagation may allow us to learn $\bm{D}$ with a low computational complexity using $O(N)$ samples.

\end{abstract}

\section{Introduction}

The concept of sparse representations has recently attracted considerable attention from various fields in which the number of measurements is limited.
Many real-world signals such as natural images are represented {\em sparsely}
in Fourier/wavelet domains; in other words, many components vanish or are negligibly
small in amplitude when the signals are represented by Fourier/wavelet bases.
This empirical property is exploited in the
signal recovery paradigm of compressed sensing (CS),
thereby enabling the recovery of
sparse signals from much fewer measurements than
those estimated by the Nyquist-Shannon sampling theorem \cite{Starck,Nyquist, Donoho,Candes}.

In signal processing techniques for exploiting sparsity, signals are generally assumed to be
described as linear combinations of a few dictionary atoms.
Therefore, the effectiveness of this approach is highly dependent on the
{choice of dictionary,
by which the objective signals appear sparse.}
A method for choosing an appropriate dictionary for sparse
representation is
{\it dictionary learning (DL)}, whereby
the dictionary is constructed through a learning process
from an available set of $P$ training samples \cite{Olshausen-Field,Rubinstein,Elad,Gleichman}.

The ambiguity of the dictionary is fatal
in signal/data analysis after learning.
Therefore, an important issue is
the estimation of the {\it sample complexity}, i.e., the sample size $P_c$ necessary for correct identification of the dictionary.
In a seminal work, Aharon et al. showed that when the training set $\bm{Y}\in \mathbb{R}^{M\times P}$ is generated by
a dictionary $\bm{D}\in \mathbb{R}^{M \times N}$ and
a sparse matrix $\bm{X} \in \mathbb{R}^{N \times P}$ (planted solution)
as $\bm{Y}=\bm{D}\bm{X}$,
one can perfectly learn these
if $P > P_{\rm c}=(k+1) {}_N C_k$ and $k$ is sufficiently small, where $k$ is the number of
non-zero elements in each column of $\bm{X}$ \cite{Aharon}.
Unfortunately, this bound becomes exponentially large in $N$ for $k \sim O(N)$, which motivates us
to improve the estimation.
A recent study has shown that 
almost all dictionaries under the uniform measure are 
learnable with $P_c\sim O(NM)$ samples when $k\ln M\sim O(\sqrt{N})$ \cite{Vainsencher2011}.
However, the fact that the number of unknown
variables $MN + NP$ and known variables $MP$ are balanced with each 
other at $P\sim O(N)$ when $M\sim O(N)$ implies the possibility of DL 
with $O(N)$ training samples.

To answer this question, in this study, we evaluate the sample complexity of the
{\rm optimal} learning scheme defined for a given probabilistic model of dictionary learning.
In a previous study, the authors assessed the sample complexity for a naive learning
scheme: $\mathop{\rm min}_{\bm{D},\bm{X}} ||\bm{Y}-\bm{D}\bm{X}||^2$ subj. to $||\bm{X}||_0 \le NP\rho$ $(0 < \rho <1)$,
where 
$||\bm{A}||$ indicates the 
Frobenius norm of $\bm{A}$, and
$||\bm{X}||_0$ is the number of non-zero elements in $\bm{X}$
and $\bm{D}$ is enforced to be normalized appropriately. 
They 
used the replica
method of statistical mechanics and found that
$P_{\rm c} \sim O(N)$ holds when $\alpha=M/N$ is greater than a certain critical value
$\alpha_{\rm naive}(\rho) > \rho$ \cite{SK}.  However, the smallest
possible $P_{\rm c}$ that can be obtained for  $\alpha < \alpha_{\rm naive} (\rho)$ has not been clarified thus far.
In this study, we show that $P_{\rm c} \sim O(N)$ holds in the entire region of $\alpha > \rho$
for the optimal learning scheme.




\section{Problem setup}
Let us suppose the following scenario of
dictionary learning. Planted solutions,
an $M \times N$ dictionary matrix $\bm{D}\in \mathbb{R}^{M\times N}$ and
an $N \times P$ sparse matrix $\bm{X} \in \mathbb{R}^{N \times P}$, are
independently generated from prior distributions,
$P(\bm{D})$ and $P_\rho(\bm{X})$, respectively,
where $P(\bm{D})$ is the uniform distribution 
over an
appropriate support and
\begin{align}
P_\rho(\bm{X})&=\prod_{i,l}P_\rho(X_{il})
\!=\!\prod_{i,l}\!\Big\{\!(1-\rho)\delta(X_{il})\!+\!
\rho f(X_{il})
\!\Big\}.
\label{eq:P_X}
\end{align}
The rate of non-zero elements in $\bm{X}$ is given by 
$\rho\in[0,1]$,
and the 
distribution function $f(X)$ does not
have a finite mass probability at the origin.
The set of training samples $\bm{Y}\in\mathbb{R}^{M\times P}$,
whose column vector corresponds to a training sample,
is assumed to be given by
the planted solutions as
\begin{align}
\bm{Y}=\frac{1}{\sqrt{N}}\bm{D} \bm{X},
\label{example}
\end{align}
where $1\slash\sqrt{N}$ is introduced for convenience
in taking the large-system limit.
A learner is required to infer 
$\bm{D}$ and $\bm{X}$ from $\bm{Y}$. 

We impose the normalization constraint of
$\sum_\mu D_{\mu i}^2=||\bm{D}_i||^2=M$
for each column $i=1\cdots,N$ to avoid the
ambiguity of the product $\bm{DX}=\bm{DA}^{-1}\bm{AX}$ for diagonal
matrices of positive diagonal entries $\bm{A}$. In addition, we
introduce two other constraints that i) the values of
$\sum_{\mu=1}^MD_{\mu i}$
are set to be positive and ii) columns of $\bm{D}$ are lined up in the
descending order of the absolute value of
$\sum_{\mu=1}^MD_{\mu i}$ so that
ambiguities of simultaneous permutation and/or multiplication
of same signs for columns in $\bm{D}$ and rows in $\bm{X}$ are removed\footnote{
One could choose different constraints as long as the trivial ambiguities
of the column order and the signs are resolved.}.
In the following, the uniform prior $P(\bm{D})$ is assumed to be
defined on the support that satisfies all of these constraints.

Our aim is to evaluate the {\em minimum} value of
the sample size $P$ required for perfectly identifying
$\bm{D}$ and {$\bm{X}$}.

\section{Bayesian optimal learning}

For mathematical formulation of our problem, let us denote the estimates of
$\bm{D}$ and $\bm{X}$ yielded by an {\em arbitrary} learning
scheme as $\hat{\bm{D}}(\bm{Y})$ and $\hat{\bm{X}}(\bm{Y})$. We evaluate the efficiency of the scheme using 
the mean squared errors (per element),
\begin{align}
&{\rm MSE}_D(\hat{\bm{D}}(\cdot))\!=\!\frac{1}{NM}\!\sum_{\bm{Y},\bm{D},\bm{X}} \!P_\rho(\!\bm{D},\!\bm{X},\!\bm{Y}\!)
||\bm{D}\!-\!\hat{\bm{D}}(\bm{Y})||^2
\label{direction_cosine}\\
&{\rm MSE}_X(\hat{\bm{X}}(\cdot))\!=\!\frac{1}{NP}\!\sum_{\bm{Y},\bm{D},\bm{X}} \!P_\rho(\!\bm{D},\!\bm{X},\!\bm{Y}\!)
||\bm{X}\!-\!\hat{\bm{X}}(\bm{Y})||^2
\label{MSE_X}
\end{align}
where $\bm{A}\cdot \bm{B} =\sum_{i,j} A_{ij}B_{ij}$ represents
the inner product between two matrices of the same dimension $\bm{A}$ and $\bm{B}$.
We impose the normalization constraint
$\sum_{\mu=1}^M(\hat{\bm{D}}(\bm{Y}))_{\mu i}^2 =M$
for each column index $i=1,2,\ldots, N$
in order to avoid the ambiguity of the product $\hat{\bm{D}}(\bm{Y})\hat{\bm{X}}(\hat{Y})
=\hat{\bm{D}}(\bm{Y}) \bm{A}^{-1} \bm{A} \hat{\bm{X}}({\bm{Y}})$
for an arbitrary invertible diagonal matrix $\bm{A}$.
The joint distribution of $\bm{D}$, $\bm{X}$, and $\bm{Y}$ is given by
\begin{align}
P_\rho(\bm{D},\bm{X},\bm{Y})=\delta(\bm{Y}-\frac{1}{\sqrt{N}}\bm{D}\bm{X})
P(\bm{D})P_\rho(\bm{X}).
\end{align}
The perfect identification of $\bm{D}$ and $\bm{X}$
can be characterized by ${\rm MSE}_D={\rm MSE}_X=0$.
The following theorem offers a useful basis for answering our question. \\

\begin{figure}
\begin{center}
\includegraphics[width=\figwidth]{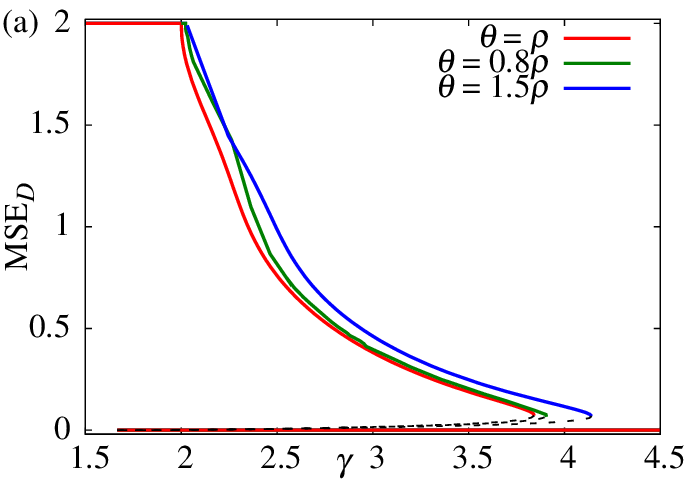}
\includegraphics[width=\figwidth]{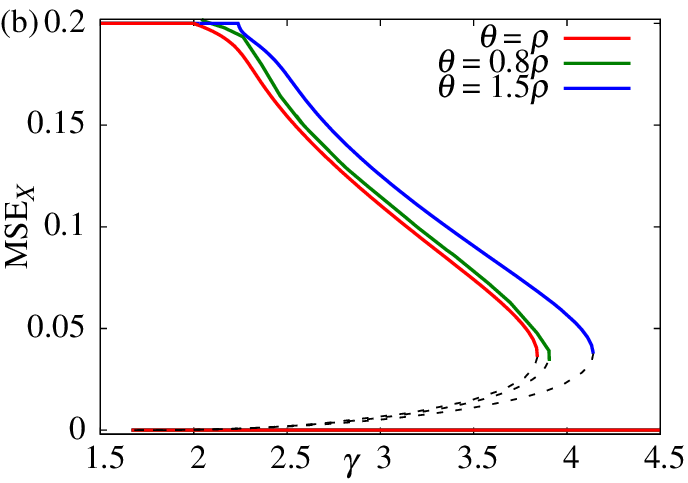}
\end{center}
\caption{$\gamma$-dependence of
(a) ${\rm MSE}_D$ and (b) ${\rm MSE}_X$ at $\alpha=0.5$, $\rho=0.2$.
{
Plots for $\theta=\rho$, $0.8\rho$, and $1.5\rho$
show the optimality of the correct parameter choice $\theta=\rho$.
Broken curves represent locally unstable branches, which
are thermodynamically irrelevant. 
}
}
\label{fig:mX_NL2}
\end{figure}

\begin{theorem}
\label{theorem1}
For an arbitrary learning scheme, (\ref{direction_cosine}) and
(\ref{MSE_X}) are bounded from below as
\begin{align}
&{\rm MSE}_D(\hat{\bm{D}}(\cdot))  \ge 2-2
\sum_{\bm{Y}}P_{\rho}(\bm{Y}) 
 \left (\frac{1}{N}\sum_{i=1}^N \frac{||(\langle \bm{D} \rangle_\rho)_i||}{\sqrt{M}} \right )
\label{bound}\\
&{\rm MSE}_X(\hat{\bm{X}}(\cdot))\ge \sum_{\bm{Y}}\!\
P_\rho(\bm{Y})\!\!\left(
\frac{\left\langle
 \bm{X}\cdot\bm{X}\right\rangle_{\rho}}{NP}\!\!-\frac{\langle \bm{X}\rangle_{\rho}\!\cdot\!\langle \bm{X}\rangle_{\rho}}{NP} \right),
\label{bound_X}
\end{align}
where $P_\rho(\bm{Y})=\sum_{\bm{D},\bm{X}}P_\rho(\bm{D},\bm{X},\bm{Y})$, and
$\langle\cdot\rangle_\rho$ denotes the average over
$\bm{D}$ and $\bm{X}$ according to the posterior distribution of $\bm{D}$ and $\bm{X}$ under a given $\bm{Y}$,
$P_\rho(\bm{D},\bm{X}|\bm{Y})=P_\rho(\bm{D},\bm{X},\bm{Y})\slash P_\rho(\bm{Y})$.
The equalities hold when the estimates satisfy
\begin{align}
(\hat{\bm{D}}^{\rm opt}\!(\bm{Y}))_i=\sqrt{M} \frac{(\left \langle  \bm{D} \right\rangle_{\rho})_i}{||(\left \langle\bm{D}\right\rangle_{\rho})_i||},~~
\hat{\bm{X}}^{\rm opt}\!(\bm{Y})=\left \langle \bm{X}\right\rangle_{\rho},
\label{Bayes_opt}
\end{align}
where $(\bm{A})_i$ denotes the $i$-th column vector of matrix $\bm{A}$.
We refer to (\ref{Bayes_opt}) as the Bayesian optimal learning scheme {\cite{Iba}}.
\end{theorem}

\noindent {\bf Proof:}
By applying the Cauchy-Shwartz inequality and the minimization of the quadratic function
to ${\rm MSE}_D$ and ${\rm MSE}_X$, respectively,
one can obtain (\ref{bound})--(\ref{Bayes_opt})
after inserting the expression
\begin{align}
\nonumber
\sum_{\bm{D},\bm{X}}x P\!\!_\rho(\bm{D}\!,\bm{X}\!,\bm{Y})
\!&=\!P\!\!_\rho(\bm{Y})\!\!\sum_{\bm{D},\bm{X}}\!\!x
 P\!\!_\rho(\bm{D}\!,\bm{X}|\bm{Y}\!)\!=\!P\!\!_\rho(\bm{Y})\!
\left\langle x\right\rangle_{\rho}
\end{align}
for $x=\bm{D}$ and $\bm{X}$ into (\ref{direction_cosine}).
\hfill $\Box$\\

This theorem guarantees that when the setup of dictionary learning is
characterized by $P(\bm{D})$ and $P_\rho(\bm{X})$,
the estimates of (\ref{Bayes_opt}) offer the {\em best possible
learning performance} in the sense that (\ref{direction_cosine}) and (\ref{MSE_X})
are minimized.
As the perfect identification of $\bm{D}$ and $\bm{X}$ is characterized by
${\rm MSE}_D={\rm MSE}_X=0$,
our purpose is fulfilled by analyzing the
performance of the Bayesian optimal learning scheme of (\ref{Bayes_opt}).

\section{Analysis}

For simplicity of calculation, let us set  $f(X_{il})$ as the Gaussian distribution
with mean 0 and variance $\sigma_X^2$,
and $\sigma_X^2$ is set to unity for all numerical calculations later on.
For generality, we consider cases in which the sparsity assumed by the learner, denoted as $\theta$, can differ from the actual value $\rho$.
When $\theta\neq\rho$,
the estimates are given by $(\hat{\bm{D}}(\bm{Y}))_i=\sqrt{M} (\langle\bm{D}\rangle_\theta)_i \slash ||(\langle \bm{D} \rangle_\theta)_i||$ and
$\hat{\bm{X}}(\bm{Y})=\langle\bm{X}\rangle_\theta$
instead of (\ref{Bayes_opt}).
To evaluate ${\rm MSE}_D$ and ${\rm MSE}_X$, we need to evaluate
macroscopic quantities
\begin{align}
q_D&\!=\!\frac{1}{MN}[\langle\bm{D}\rangle_\theta\!\!\cdot\!\langle\bm{D}\rangle_{\theta}]_{Y},~~
m_D\!=\!\frac{1}{MN}[\langle\bm{D}\rangle_{\theta}\!\!\cdot\!\langle\bm{D}\rangle_{\rho}]_{Y}\label{eq:q_Dm_D}\\
Q_X&=\frac{1}{NP}[\langle\bm{X}\!\cdot\!\bm{X}\rangle_{\theta}]_{Y}\label{eq:Q_X}\\
q_X&\!=\!\frac{1}{NP}[\langle\bm{X}\rangle_{\theta}\!\!\cdot\!\langle\bm{X}\rangle_{\theta}]_{Y},~~
m_X\!=\!\frac{1}{NP}[\langle\bm{X}\rangle_{\theta}\!\!\cdot\!\langle\bm{X}\rangle_{\rho}]_{Y},\label{eq:q_Xm_X}
\end{align}
where $[\cdot]_Y=\sum_{\bm{Y}}P_\rho(\bm{Y})(\cdot)$.
Note that (\ref{eq:q_Dm_D})--(\ref{eq:q_Xm_X}) yield 
${\rm MSE}_D\simeq 2-2 m_D/\sqrt{q_D}$\footnote{
{
Naive computation requires us to assess a column-wise overlap 
$C_{D,i}=M^{-1/2} [  (\langle \bm{D} \rangle_\rho)_i \!\! \cdot \!\! (\langle \bm{D} \rangle_\theta)_i
\slash \sqrt{(\langle \bm{D} \rangle_\theta)_i \!\! \cdot \!\! (\langle \bm{D} \rangle_\theta)_i} ]_{Y}$ 
for each column index $i=1,2,\ldots,N$. However, the law of large numbers and 
the statistical uniformity 
allow the simplification of $C_{D,i} \to m_D/\sqrt{q_D}$ as $M=\alpha N $ tends to infinity. }
}
and ${\rm MSE}_X=\rho\sigma_X^2+q_X-2m_X$.

Unfortunately, evaluating these is intrinsically difficult
because it generally requires averaging the quantity
\begin{eqnarray}
&&
\frac{\sum_{\!\bm{D}^1,\!\bm{X}^1,\!\bm{D}^2,\!\bm{X}^2} P_\theta (\bm{Y},\!\bm{D}^1,\!\bm{X}^1)
P_\theta (\bm{Y},\!\bm{D}^2,\!\bm{X}^2) (\bm{D}^1\cdot \bm{D}^2) }
{\sum_{\!\bm{D}^1,\!\bm{X}^1,\!\bm{D}^2,\!\bm{X}^2}P_\theta (\bm{Y},\!\bm{D}^1,\!\bm{X}^1)
P_\theta (\bm{Y},\!\bm{D}^2,\!\bm{X}^2)} \cr
&&(=\langle\bm{D}\rangle_\theta\!\!\cdot\!\langle\bm{D}\rangle_{\theta}),
\label{denominator}
\end{eqnarray}
which includes summations over exponentially many terms in the denominator,
with respect to $\bm{Y}$.
One promising approach for avoiding this difficulty involves multiplying $P_\theta^n(\bm{Y})=(\sum_{\bm{X},\bm{D}} P_\theta (\bm{Y},\bm{D},\bm{X}))^n$
$(n ={2,3},\ldots \in \mathbb{N})$ inside the operation of $[\cdot ]_Y$ for
canceling the denominator of (\ref{denominator}), which makes the evaluation of a modified average
\begin{eqnarray}
q_D(n) =\frac{1}{MN}
\frac{[ P_\theta^n(\bm{Y}) \langle\bm{D}\rangle_\theta\!\!\cdot\!\langle\bm{D}\rangle_{\theta}]_{Y}}
{[ P_\theta^n(\bm{Y})]_{Y}}
\end{eqnarray}
feasible via the saddle point assessment of
$[P^n_\theta (\bm{Y})]_{Y}$
for $N,M,P \to \infty$, keeping $\alpha=M/N$ and $\gamma=P/N$ as $O(1)$.
Furthermore, the resulting expression is likely to hold for $n \in \mathbb{R}$ as well.
Therefore, we evaluate $q_D$ using the formula $q_D=\lim_{n \to 0}q_D(n)$
with the expression, and similarly, for $m_D$, $Q_X$, $q_X$, and $m_X$.
This procedure is often termed the {\em replica method} \cite{beyond,Nishimori}.
Under the replica symmetric ansatz, which assumes that the dominant
saddle point in the evaluation is invariant under any permutation of
replica indices $a=1,2,\ldots, n$, the assessment is reduced to evaluating the extremum of
the free entropy ({density}) function
\begin{align}
\nonumber
\phi
&=
\gamma\Big(\frac{\hat{Q}_XQ_X+\hat{q}_Xq_X}{2}-\hat{m}_Xm_X+\langle\langle\ln\Xi_X\rangle\rangle\Big)\\
\nonumber
&+\frac{\alpha}{2}\!\Big(\hat{Q}_D\!+\!\hat{q}_Dq_D\!-\!2\hat{m}_Dm_D\!-\!\ln(\hat{Q}_D\!+\!\hat{q}_D)\!+\!\frac{\hat{q}_D\!+\!\hat{m}_D^2}{\hat{Q}_D\!+\!\hat{q}_D}\Big)\\
&-\!\frac{\alpha\gamma}{2}\!\Big\{\!\frac{q_Dq_X\!-\!2m_Dm_X\!+\!\rho\sigma_X^2\!}{Q_X-q_Dq_X}\!+\!\ln(Q_X\!-\!q_Dq_X)\Big\}\!
,
\end{align}
where $\hat\sigma_X=1+(\hat{Q}_X+\hat q_X)\sigma_X^{2}$,
\begin{align}
\nonumber
\Xi_X\!\!&=\!(1\!-\!\theta)\!+\!\frac{\theta}{\sqrt{\hat\sigma_X}}
\exp\!\Big(\frac{\sigma_X^2 (\sqrt{\hat q}_Xz+\hat
 m_XX^0)^2}{2\hat\sigma_X}\!\Big) \\
 &\equiv(1-\!\theta\!)+\!\Xi_X^{+},
\end{align}
and $\langle\langle\cdot\rangle\rangle$ denotes the average over $X$ and $z$,
which are distributed according to $P_\rho(X)$
and a Gaussian distribution with mean zero and variance 1, respectively.
{
The extremized value\footnote{When multiple extrema exist, the maximum
value among them should be chosen as long as no consistency condition is violated.}
of $\phi$, $\phi^*$, 
is related to the average log-likelihood (density) of $\bm{Y}$ 
as $N^{-2}\sum_{\bm{Y}}P_\rho(\bm{Y})\ln P_\theta(\bm{Y})
=\lim_{n\to 0} (\partial /\partial n)
\left \{N^{-2} \ln [P_\theta^n (\bm{Y}) ]_Y \right \}
=\phi^*+{\rm constant}.$}

In \Fref{fig:mX_NL2},
(a) ${\rm MSE}_D$ and (b) ${\rm MSE}_X$
for $\theta=\rho=0.2$ are plotted versus $\gamma$
together with those for
$\theta=0.8 \rho$ and $\theta=1.5\rho$.
At $\theta=\rho$, ${\rm MSE}_D$ and ${\rm MSE}_X$ 
{of thermodynamically relevant branches} 
have minimum values in the entire $\gamma$ region,
{while a branch of solution characterized by 
${\rm MSE}_D={\rm MSE}_X=0$ is shared by the three parameter sets. 
This supports the optimality of the correct parameter choice
of $\theta=\rho$, and therefore, 
we hereafter focus our analysis on this case
to estimate
the minimum value of $\gamma$ for the perfect learning, 
${\rm MSE}_D={\rm MSE}_X=0$.
At $\theta=\rho$, the relationships
$m_D=q_D$, $m_X=q_X$, and $Q_X=\rho$ hold
from (\ref{eq:q_Dm_D})--(\ref{eq:q_Xm_X}),
and the extremum problem is reduced to
\begin{align}
q_D&\!=\!\frac{\hat q_D}{1+\hat q_D},~~
q_X\!=\!\left<\left<\!\!\left(\frac{\Xi_X^+}{\Xi_X}\frac{\sqrt{\hat q_X}z+\hat q_XX^0}{\hat\sigma_X^2}\!\right)^2\right>\right>,\label{eq:q_X}
\end{align}
where
$\hat q_D$ and $\hat q_X$ are given by
\begin{align}
\hat{q}_X=\frac{\alpha q_D}{\rho\sigma_X^2-q_Dq_X},~~~
\hat{q}_D=\frac{\gamma q_X}{\rho\sigma_X^2-q_Dq_X}.
\end{align}
The other
variables are provided as $\hat Q_D=1$, $\hat Q_X=0$,
$\hat m_X=\hat q_X$, and $\hat m_D = \hat q_D$.

\section{Results}

\subsection{Actual solutions}

\begin{figure}
\begin{center}
\includegraphics[width=\figwidth]{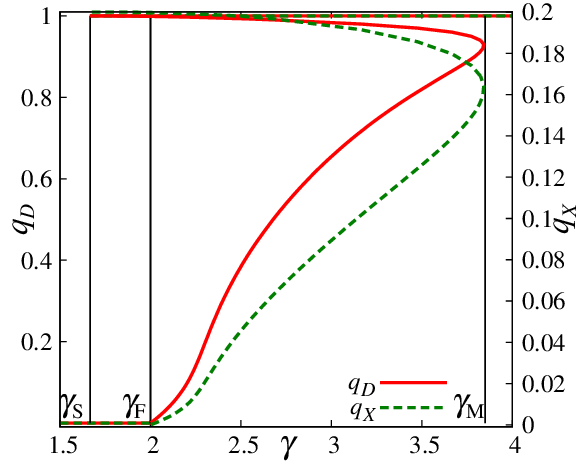}
\end{center}
\caption{$\gamma$-dependence of $q_D$ (left axis) and $q_X$ (right axis)
for $\alpha=0.5$ and $\rho=0.2$.
}
\label{fig:alpha05rho031}
\end{figure}


\Fref{fig:alpha05rho031} plots
$q_D$ and $q_X$ versus $\gamma$
for $\alpha=0.5$ and $\rho=0.2$.
As shown in the figure, the solutions of $q_D$ and $q_X$ given by (\ref{eq:q_X})
are classified into three types:
$q_D=1,~q_X=\rho {\sigma_X^2}$, $q_D=q_X=0$, and $0 < q_D < 1,~0<q_X<\rho {\sigma_X^2}$.
The first one yields ${\rm MSE}_D={\rm MSE}_X=0$,
indicating the correct identification of $\bm{D}$ and $\bm{X}$,
and hence, we name it the {\it success solution}.
The second one is referred to as the {\it failure solution}
because it yields ${\rm MSE}_D=2$ and ${\rm MSE}_X=\rho \sigma_X^2$,
which indicates complete failure of the learning of $\bm{D}$ and $\bm{X}$.
The third one yields finite ${\rm MSE}_D$ and ${\rm MSE}_X$, $0<{\rm MSE}_D<2,~0<{\rm MSE}_X<\rho \sigma_X^2$,
and we term it the {\it middle solution}.

\subsubsection{Success solution}
When the expression
\begin{align}
\delta\Big(\bm{Y}\!\!-\!\frac{\bm{DX}}{\sqrt{N}}\!\Big)\!=\!\lim_{\tau\to
 +0}\!\!\Big(\!\frac{1}{\sqrt{2\pi\tau}}\!\Big)^{MP}\!\!\!\!\!\!\!\!\exp\!\!\Big(\!\!-\!\frac{||\bm{Y}\!\!\!-\!\frac{1}{\sqrt{N}}\bm{DX}||^2}{2\tau}\!\Big),
\label{eq:delta}
\end{align}
is used,
the success solution of $q_D$ and $q_X$
behaves as $(\rho\sigma_X^2-q_X)\slash\tau=\chi_X$ and $(1-q_D)\slash\tau=\chi_D$
while $\hat{q}_X$ and $\hat{q}_D$ scale as
$\hat q_X=\hat\theta_X\slash\tau$ and $\hat q_D=\hat\theta_D\slash\tau$.
By substituting them into the equations of $q_D$ and
$q_X$, they are given by
\begin{align}
\chi_X=\frac{\rho\gamma}{g},~~~\chi_D=\frac{\alpha}{\rho\sigma_X^2g},~~~
\hat\theta_X=\frac{\rho}{\chi_X},~~~\hat\theta_D=\frac{1}{\chi_D},
\end{align}
where $g=(\alpha-\rho)\gamma-\alpha$.
$\chi_X$ and $\chi_D$ must be positive by
definition, and hence, the success solution exists for
\begin{align}
\gamma > \frac{\alpha}{\alpha-\rho}\equiv\gamma_{\rm S}
\end{align}
only when $\alpha>\rho$.

\subsubsection{Failure solution}
The failure solution $q_D=q_X=0$ appears at $0\leq\gamma<\gamma_{\rm F}$
as a locally stable solution.
When $q_D$ and $q_X$ are sufficiently small, they are expressed as
\begin{align}
q_X&=\rho\sigma_X^2\alpha q_D+O(q^2),~~~q_D=\frac{\gamma q_X}{\rho\sigma_X^2}+O(q^2),
\end{align}
where $O(q^2)$ denotes the
higher-order terms over second-order with respect to $q_D$ and $q_X$.
These expressions indicate that when
\begin{align}
\gamma > \alpha^{-1}\equiv\gamma_{\rm F},
\end{align}
the local stability of $q_D=q_X=0$ is lost. 
As shown in \Fref{fig:alpha05rho031},
the failure solution vanishes at $\gamma_{\rm F}=2.0$
for $\alpha=0.5$.

\subsubsection{Middle solution}

\begin{figure}
\begin{center}
\includegraphics[width=\figwidth]{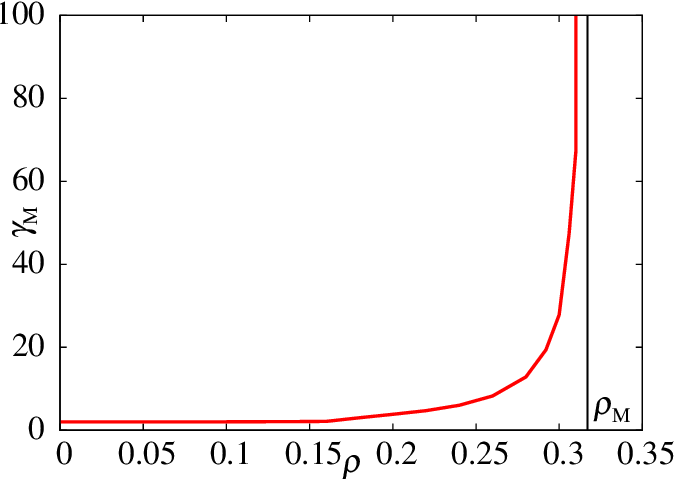}
\end{center}
\caption{$\gamma_{\rm M}$ versus $\rho$ for $\alpha=0.5$. }
\label{fig:gamma_c}
\end{figure}

We define $\gamma_{\rm M}$ over which the
middle solution with $0<q_D<1$ and $0<q_X<\rho \sigma_X^2$ disappears,
denoted as a vertical line in \Fref{fig:alpha05rho031}, 
{which is provided as $\gamma_{\rm M}=3.841\ldots$
for the parameter choice of $(\alpha,\rho)=(0.5,0.2)$}.
The value of $\gamma_{\rm M}$ depends on {$(\alpha,\rho)$},
as shown in \Fref{fig:gamma_c}. 
This figure indicates that $\gamma_{\rm M}$ diverges at $\rho_{\rm M} =0.317\ldots$ for $\alpha=0.5$.
The relation between $\rho_{\rm M}$ and $\alpha$, denoted as $\rho_{\rm M}(\alpha)$ (or $\alpha_{\rm M}(\rho)$),
generally accords with the critical condition that
belief propagation (BP)-based signal recovery using the correct prior
starts to be involved with multiple fixed points for the signal reconstruction problem of compressed sensing \cite{Krzakala2012} in which
the correct dictionary $\bm{D}$ is provided in advance.

BP is also a potential algorithm for practically
achieving the learning performance predicted by the current analysis because it is known that macroscopic behavior {\em theoretically} analyzed by the replica method can be confirmed {\em experimentally} for single instances by BP for many other systems \cite{Krzakala2012,TAP,Kabashima2003}.
The fact that only the success solution exists for $\gamma > \gamma_{\rm M}$ implies that
one may be able to perfectly identify the correct dictionary $\bm{D}$ with a
computational cost of {\em polynomial} order in $N$ utilizing BP, without being
trapped by other locally stable solutions, for
$\alpha> \alpha_{\rm M}(\rho)$.

\subsection{Free entropy density}

\begin{figure}
\begin{center}
\includegraphics[width=\figwidth]{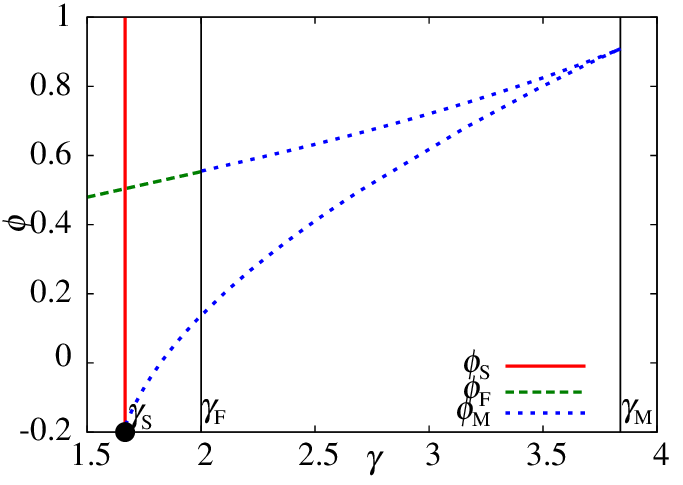}
\end{center}
\caption{$\gamma$-dependence of $\phi$
for $\alpha=0.5$ and $\rho=0.2$.
$\phi_{\rm S}$ diverges positively for $\gamma > \gamma_{\rm S}=\alpha/(\alpha-\rho)=1.666\ldots$.
}
\label{fig:lnZ}
\end{figure}

There are three extrema of the free entropy (density), $\phi_{\rm S}$,
$\phi_{\rm F}$, and $\phi_{\rm M}$,
corresponding to the success solution, failure solution,
and middle solution, respectively.
Among them, the thermodynamically dominant solution
that provides the correct evaluations of $q_D$ and $q_X$ is
the one for which the value of free entropy is the largest.
\Fref{fig:lnZ} plots $\phi_{\rm S}$,
$\phi_{\rm F}$, and $\phi_{\rm M}$ versus $\gamma$
for  $\alpha=0.5$, $\rho=0.2$, where $\gamma_{\rm S}=1.666\ldots$
and $\gamma_{\rm F}=2.0$.
In particular, functional forms of $\phi_{\rm S}$ and $\phi_{\rm F}$ are given by
\begin{align}
\nonumber
\phi_{\rm S}\!&=\lim_{\tau\to +0}\!\frac{1}{2}\Big[g\!\left\{\ln(\frac{g}{\tau}
)\!-\!1\right\}\!-\!\alpha\gamma\ln(\alpha\gamma)\!+\!\alpha
\!\left\{1\!-\!\ln\left(\frac{\rho\sigma_X^2}{\alpha}\right)\!\right\} \\
&\hspace{1.0cm}\!+\!\gamma\rho
 (\ln\gamma-\ln\sigma_X^2)\Big]\!-\!\gamma H(\rho) \label{phiS}\\
\phi_{\rm F}&=\frac{1}{2}\{-\alpha\gamma(1+\log\rho\sigma_X^2)+\alpha\},
\end{align}
where $\tau\to +0$ originates from
the expression of (\ref{eq:delta}) and
$H(\rho)=-(1-\rho)\log(1-\rho)-\rho\log(\rho)$.
Further, (\ref{phiS}) shows that $\phi_{\rm S}$ diverges positively
for $g=(\alpha-\rho)\gamma -\alpha >0$,
which guarantees that the success solution is
always thermodynamically dominant for $\gamma > \gamma_{\rm S}=\alpha/(\alpha-\rho)$
as $\phi$ of other solutions is kept finite.
This leads to the conclusion that the sample complexity of
the Bayesian optimal learning is $P_{\rm c}=N \gamma_{\rm S}$,
which is guaranteed as $O(N)$ as long as $\alpha > \rho$.
This is the main consequence of the present study.

\Fref{fig:alpha-rho} plots the phase diagram in the $\alpha-\rho$ plane.
The union of the regions (I) and (II) represents the condition that
the sample complexity $P_{\rm c}$ is $O(N)$,
while the full curve of the upper boundary of (II) denotes
$\alpha_{\rm M}(\rho)$ above which BP is expected to work
as an efficient learning algorithm.
Dictionary learning is impossible in the region of (III).
The critical condition $\alpha_{\rm {naive}}(\rho)$ above
which the naive learning scheme of
\cite{SK} can perfectly identify the planted solution by $O(N)$ samples
is drawn as the dashed curve for comparison.
The considerable difference between $\alpha_{\rm naive}(\rho)$ and $\rho$
(or even $\alpha_{\rm M}(\rho)$) indicates the significance of
using adequate knowledge of probabilistic models in
dictionary learning.

\begin{figure}
\begin{center}
\includegraphics[width=\figwidth]{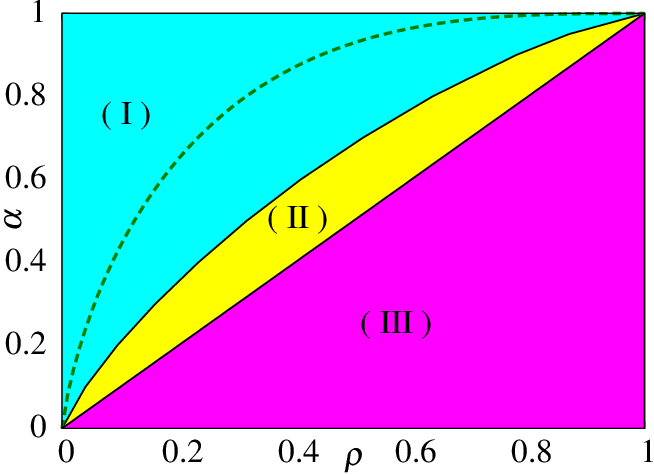}
\end{center}
\caption{Phase diagram on $\alpha-\rho$ plane.
{The dashed curve in the 
area of (I) is the result of \cite{SK}.}}
\label{fig:alpha-rho}
\end{figure}

\section{Summary}
In summary, we assessed the minimum sample size
required for perfectly identifying a planted solution in dictionary learning (DL).
For this assessment, we derived the optimal learning scheme defined for
a given probabilistic model of DL following the framework of Bayesian inference.
Unfortunately, actually evaluating the performance of the Bayesian optimal
learning scheme involves an intrinsic technical difficulty.
For resolving this difficulty, we resorted to the replica method of
statistical mechanics, and we showed that the sample complexity can be
reduced to $O(N)$ as long as the compression rate $\alpha $ is greater than
the density  $\rho$ of non-zero elements of the sparse matrix.
This indicates that the performance of a naive learning scheme examined in
a previous 
study \cite{SK} can be improved significantly by
utilizing the knowledge of adequate probabilistic models in DL.
It was also shown that when $\alpha$ is greater than a certain
critical value $\alpha_{\rm M}(\rho)$, the macroscopic state corresponding to
perfect identification of the planted solution becomes a unique candidate
for the thermodynamically dominant state. This suggests that
one may be able to learn the planted solution with a computational complexity
of polynomial order in $N$ utilizing belief propagation for $\alpha > \alpha_{\rm M}(\rho)$. \\
{\em -- Note added: After completing this study,  the authors became aware that
\cite{Krzakala2013} presents results similar to those presented in this paper,
where an algorithm for dictionary {learning/calibration} is independently developed on the basis of belief propagation.}

%


\section*{Acknowledgment}

This work was partially supported by 
a Grant-in-Aid for JSPS Fellow No. 23--4665 (AS)
and KAKENHI Nos. 22300003 and 22300098 (YK),
and JSPS Core-to-Core Program ``Nonequilibrium
dynamics of soft matter and information''.


\begin{thebibliography}{1}
\expandafter\ifx\csname natexlab\endcsname\relax\def\natexlab#1{#1}\fi
\expandafter\ifx\csname bibnamefont\endcsname\relax
 \def\bibnamefont#1{#1}\fi
\expandafter\ifx\csname bibfnamefont\endcsname\relax
 \def\bibfnamefont#1{#1}\fi
\expandafter\ifx\csname citenamefont\endcsname\relax
 \def\citenamefont#1{#1}\fi
\expandafter\ifx\csname url\endcsname\relax
 \def\url#1{\texttt{#1}}\fi
\expandafter\ifx\csname urlprefix\endcsname\relax\def\urlprefix{URL}\fi
\providecommand{\bibinfo}[2]{#2}
\providecommand{\eprint}[2][]{\url{#2}}

\bibitem{Starck}
\bibinfo{author}{\bibfnamefont{J.-L.}~\bibnamefont{Starck}},
\bibinfo{author}{\bibfnamefont{F.}~\bibnamefont{Murtagh }},
\bibnamefont{and}
\bibinfo{author}{\bibfnamefont{J.~M.}~\bibnamefont{Fadili}},
\emph{\bibinfo{title}{Sparse Image and Signal Processing: Wavelets, Curvelets, Morphological Diversity}}
(\bibinfo{publisher}{Cambridge Univ. Press}, \bibinfo{address}{New York}, \bibinfo{year}{2010}).

\bibitem{Nyquist}
\bibinfo{author}{\bibfnamefont{H.}~\bibnamefont{Nyquist}},
\emph{\bibinfo{title}{Certain topics in telegraph transmission theory}},
\bibinfo{journal}{Trans. AIEE}~\textbf{\bibinfo{volume}{47}}~{\bibinfo{number}{(2)}},
\bibinfo{pages}{pp.~617--644} (\bibinfo{year}{1928}).




\bibitem{Donoho}
\bibinfo{author}{\bibfnamefont{D.~L.}~\bibnamefont{Donoho}},
\emph{\bibinfo{title}{Compressed sensing}},
\bibinfo{journal}{IEEE Trans. Inform. Theory}~\textbf{\bibinfo{volume}{52}}~\bibinfo{number}{(4)},
\bibinfo{pages}{pp.~1289--1306} (\bibinfo{year}{2006}).

\bibitem{Candes}
\bibinfo{author}{\bibfnamefont{E.~J.}~\bibnamefont{Cand\`{e}s}},~\bibnamefont{and}
\bibinfo{author}{\bibfnamefont{T}~\bibnamefont{Tao}},
\emph{\bibinfo{title}{Decoding by Linear Programming}},
\bibinfo{journal}{IEEE Trans. Inform. Theory}~\textbf{\bibinfo{volume}{51}}~\bibinfo{number}{(12)},
\bibinfo{pages}{pp.~4203--4215} (\bibinfo{year}{2005}).




\bibitem{Olshausen-Field}
\bibinfo{author}{\bibfnamefont{B.~A.}~\bibnamefont{Olshausen}}
 \bibnamefont{and}
 \bibinfo{author}{\bibfnamefont{D.~J.}~\bibnamefont{Field}},
 \emph{\bibinfo{title}{Sparse Coding with an Overcomplete Basis Set: A Strategy Employed by V1?}},
 \bibinfo{journal}{Vision Res.}~\textbf{\bibinfo{volume}{37}}~\bibinfo{number}{(23)},
 \bibinfo{pages}{pp.~3311--3325} (\bibinfo{year}{1997}).

\bibitem{Rubinstein}
\bibinfo{author}{\bibfnamefont{R.}~\bibnamefont{Rubinstein}},
 \bibinfo{author}{\bibfnamefont{A.~M.}~\bibnamefont{Bruckstein}}, \bibnamefont{and}
 \bibinfo{author}{\bibfnamefont{M.}~\bibnamefont{Elad}},
 \emph{\bibinfo{title}{Dictionaries for Sparse Representation Modeling}},
 \bibinfo{journal}{Proc. of IEEE}~\textbf{\bibinfo{volume}{98}}~\bibinfo{number}{(6)},
 \bibinfo{pages}{pp.~1045--1057} (\bibinfo{year}{2010}).

\bibitem{Elad}
\bibinfo{author}{\bibfnamefont{M.} \bibnamefont{Elad}},
 \emph{\bibinfo{title}{Sparse and Redundant Representations: From Theory to Applications in Signal and Image Processing}},
(\bibinfo{publisher}{Springer-Verlag}, \bibinfo{address}{New York}, \bibinfo{year}{2010}).

\bibitem{Gleichman}
\bibinfo{author}{\bibfnamefont{S.}~\bibnamefont{Gleichman}}, \bibnamefont{and}
 \bibinfo{author}{\bibfnamefont{Y.~C.}~\bibnamefont{Eldar}},
 \emph{\bibinfo{title}{Blind Compressed Sensing}},
 \bibinfo{journal}{IEEE Inform. Theory} \textbf{\bibinfo{volume}{57}},
 \bibinfo{pages}{pp. 6958--6975} (\bibinfo{year}{2011}).

\bibitem{Aharon}
\bibinfo{author}{\bibfnamefont{M.}~\bibnamefont{Aharon}},
\bibinfo{author}{\bibfnamefont{M.}~\bibnamefont{Elad}},
 \bibnamefont{and} \bibinfo{author}{\bibfnamefont{A.~M.}~\bibnamefont{Bruckstein}},
 \emph{\bibinfo{title}{On the uniqueness of overcomplete dictionaries, and a practical way to retrieve them}},
 \bibinfo{journal}{Linear Algebra and its Applications}
 \textbf{\bibinfo{volume}{416}}~\bibinfo{number}(1), \bibinfo{pages}{pp.~48--67} (\bibinfo{year}{2006}).

\bibitem{Vainsencher2011}
\bibinfo{author}{\bibfnamefont{D.}~\bibnamefont{Vainsencher}},
\bibinfo{author}{\bibfnamefont{S.}~\bibnamefont{Mannor}},
\bibnamefont{and}
\bibinfo{author}{\bibfnamefont{A.~M.}~\bibnamefont{Bruckstein}},
\emph{\bibinfo{title}{The Sample Complexity of Dictionary Learning}},
\bibinfo{journal}{Journal of Machine Learning Research}
\textbf{\bibinfo{volume}{12}}, \bibinfo{pages}{pp.~3259--3281} (\bibinfo{year}{2011}).

\bibitem{SK}
\bibinfo{author}{\bibfnamefont{A.}~\bibnamefont{Sakata}}, \bibnamefont{and}
 \bibinfo{author}{\bibfnamefont{Y.}~\bibnamefont{Kabashima}},
 \emph{\bibinfo{title}{Statistical mechanics of dictionary learning}},
 \bibinfo{note}{arXiv:1203.6178}.

\bibitem{Iba}
\bibinfo{author}{\bibfnamefont{Y.}~\bibnamefont{Iba}},
\emph{\bibinfo{title}{The Nishimori line and Bayesian statistics}},
\bibinfo{journal}{J. Phys. A: Math. Gen.} \textbf{\bibinfo{volume}{32}} 
\bibinfo{number}{(21)}, \bibinfo{page}{3875--3888} \bibinfo{year}{(1999)}.







\bibitem{beyond}
\bibinfo{author}{\bibfnamefont{M.}~\bibnamefont{M$\acute{\mbox{e}}$zard}},
 \bibinfo{author}{\bibfnamefont{G.}~\bibnamefont{Parisi}}, \bibnamefont{and}
 \bibinfo{author}{\bibfnamefont{M.~A.} \bibnamefont{Virasoro}},
 \bibinfo{title}{\emph{Spin Glass Theory and Beyond}},
 (\bibinfo{publisher}{World Sci. Pub.}, \bibinfo{year}{1987}).

\bibitem{Nishimori}
\bibinfo{author}{\bibfnamefont{H.}~\bibnamefont{Nishimori}},
 \bibinfo{title}{\emph{Statistical Physics of Spin Glasses and Information
 Processing: An Introduction}}, (\bibinfo{publisher}{Oxford Univ. Pr.},
 \bibinfo{year}{2001}).

\bibitem{Mezard-Montanari}
\bibinfo{author}{\bibfnamefont{M.}~\bibnamefont{M\'{e}zard}} \bibnamefont{and}
\bibinfo{author}{\bibfnamefont{A.}~\bibnamefont{Montanari}},
 \emph{\bibinfo{title}{Information, Physics, and Computation}},
(\bibinfo{publisher}{Oxford Univ. Press}, \bibinfo{address}{Oxford, UK}, \bibinfo{year}{2009}).

\bibitem{Krzakala2012}
\bibinfo{author}{\bibfnamefont{F.}~\bibnamefont{Krzakala}},
\bibinfo{author}{\bibfnamefont{M.}~\bibnamefont{M\'{e}zard}},
\bibinfo{author}{\bibfnamefont{F.}~\bibnamefont{Sausset}},
\bibinfo{author}{\bibfnamefont{Y.~F.}~\bibnamefont{Sun}},
\bibnamefont{and}
\bibinfo{author}{\bibfnamefont{L.}~\bibnamefont{Zdeborov\'{a}}}
\emph{\bibinfo{title}{Statistical-Physics-Based Reconstruction in Compressed Sensing}},
\bibinfo{journal}{Phys. Rev. X}
\textbf{\bibinfo{volume}{2}}, \bibinfo{pages}{pp. 021005-1--021005-18} (\bibinfo{year}{2012}).

\bibitem{TAP}
\bibinfo{author}{\bibfnamefont{D.~J.}~\bibnamefont{Thouless}},
\bibinfo{author}{\bibfnamefont{P.~W.}~\bibnamefont{Anderson}},
\bibnamefont{and}
\bibinfo{author}{\bibfnamefont{R.~G.}~\bibnamefont{Palmer}},
\emph{\bibinfo{title}{Solution of 'Solvable model of a spin glass}},
\bibinfo{journal}{Phil. Mag.}
\textbf{\bibinfo{volume}{35}}~\bibinfo{number}{(3)}, \bibinfo{pages}{pp.~ 593--601} (\bibinfo{year}{1977}).

\bibitem{Kabashima2003}
\bibinfo{author}{\bibfnamefont{K.}~\bibnamefont{Kabashima}},
\emph{\bibinfo{title}{A CDMA multiuser detection algorithm on the basis of belief propagation}},
\bibinfo{journal}{J. Phys. A}
\textbf{\bibinfo{volume}{36}}~\bibinfo{number}{(43)}, \bibinfo{pages}{pp.~11111--11121} (\bibinfo{year}{2003}).


\bibitem{Krzakala2013}
\bibinfo{author}{\bibfnamefont{F.}~\bibnamefont{Krzakala}},
\bibinfo{author}{\bibfnamefont{M.}~\bibnamefont{M\'{e}zard}},
\bibnamefont{and}
\bibinfo{author}{\bibfnamefont{L.}~\bibnamefont{Zdeborov\'{a}}},
\emph{\bibinfo{title}{Phase Diagram and Approximate Message Passing
for Blind Calibration and Dictionary Learning}}. 
\bibinfo{note}{Preprint received directly from the authors via private communication}.

\end{thebibliography}

\end{document}